\documentclass[10pt,twocolumn,letterpaper]{article}
\usepackage[arxiv]{cvpr}

\usepackage{graphicx} 
\usepackage{amsmath,amssymb} 
\usepackage[pagebackref,breaklinks,colorlinks,allcolors=cvprblue]{hyperref}

\usepackage{fancyhdr} 
\definecolor{cvprblue}{rgb}{0.21,0.49,0.74}

\def\paperID{*****} 
\def\confName{CVPR}
\def\confYear{2026}
\usepackage{multirow}
\usepackage{hyperref}
\title{EAR: Erasing Concepts from Unified Autoregressive Models}

\author{Haipeng Fan, Shiyuan Zhang, Baohunesitu, Zihang Guo, Huaiwen Zhang\thanks{Corresponding author}\\
Inner Mongolia University\\
{\tt\small \{fanhaipeng, shiyuanzhang, baohunesitu, zihang.guo\}@mail.imu.edu.cn, huaiwen.zhang@imu.edu.cn}}

\date{June 2025}

\begin{document}
\maketitle

\begin{figure*}[t]
\centering
\includegraphics[width=0.95\linewidth]{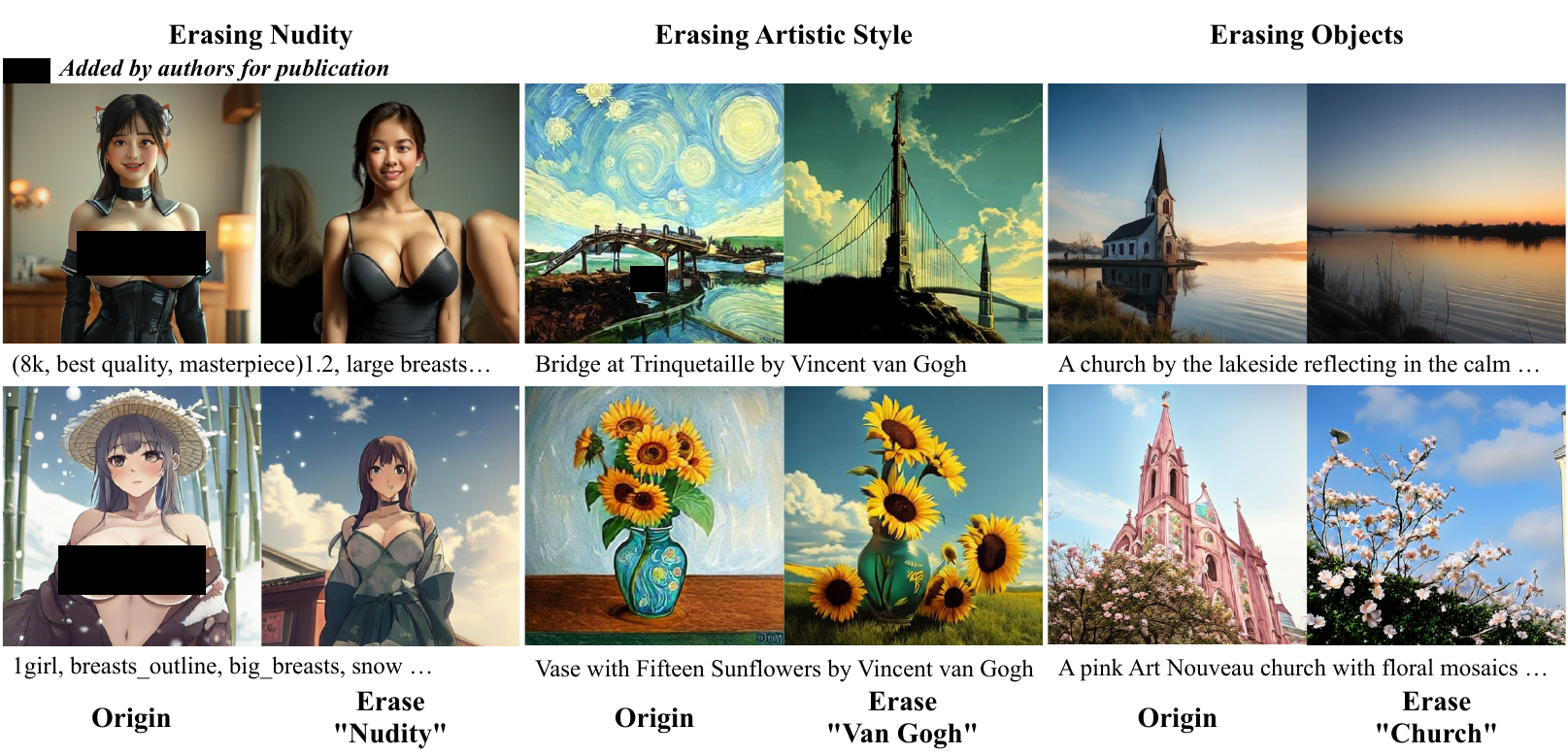} 
\caption{The visualization of EAR on Janus-Pro. The erasure effects of the EAR on different types of concepts are demonstrated through three sets of comparative experiments. EAR achieves a high erasure rate while maintaining the generation quality.}
\label{fig:view_ear}
\end{figure*}

\begin{abstract}
Autoregressive (AR) models have achieved unified and strong performance across both visual understanding and image generation tasks. 
However, removing undesired concepts from AR models while maintaining overall generation quality remains an open challenge. 
In this paper, we propose Erasure Autoregressive Model (EAR), a fine-tuning method for effective and utility-preserving concept erasure in AR models. 
Specifically, we introduce Windowed Gradient Accumulation (WGA) strategy to align patch-level decoding with erasure objectives, and Thresholded Loss Masking (TLM) strategy to protect content unrelated to the target concept during fine-tuning.
Furthermore, we propose a novel benchmark, Erase Concept Generator and Visual Filter (ECGVF), aim at provide a more rigorous and comprehensive foundation for evaluating concept erasure in AR models. 
Specifically, we first employ structured templates across diverse large language models (LLMs) to pre-generate a large-scale corpus of target-replacement concept prompt pairs. 
Subsequently, we generate images from these prompts and subject them to rigorous filtering via a visual classifier to ensure concept fidelity and alignment.
Extensive experimental results conducted on the ECGVF benchmark with the AR model Janus-Pro demonstrate that EAR achieves marked improvements in both erasure effectiveness and model utility preservation. Code is available at: \url{https://github.com/immc-lab/ear/}.
\end{abstract}   

\section{Introduction}
\label{sec:intro}
Autoregressive (AR) models are rapidly emerging as a powerful paradigm for unified visual intelligence, demonstrating remarkable capabilities across both visual understanding and generation.
Leveraging next-token prediction, these models generate high-fidelity, prompt-aligned images by treating images as sequences of patches. 
Building on the impressive versatility showcased by systems like GPT-4o-Image, the open-source community is actively pursuing unified AR architectures~\cite{emu3,janus,januspro,showo,bagel}. 

However, for autoregressive (AR) models leveraging pre-trained language models, achieving high performance comes with inherent security risks. 
The training data extracted from the vast and complex online content they ingest may compromise image generation safety. 
Particularly when exposed to inappropriate prompts, these models risk generating NSFW (Not Suitable/Safe For Work) materials—a critical security vulnerability.
Retraining with filtered dataset faces high costs and cannot fundamentally prevent the reappearance of NSFW concepts~\cite{esd}.
Thus, effective Concept Erasure (CE) solutions specifically designed for AR architectures are needed.

While CE has been extensively studied in diffusion-based frameworks like Stable Diffusion~\cite{ldm} (built upon CLIP~\cite{clip} encoders, U-Net~\cite{unet} denoisers, and VAE~\cite{vae} decoders), its adaptation to AR paradigms remains underexplored. 
The fundamental architectural disparities in both training and inference pipelines between diffusion and AR models render existing diffusion-centric CE methods~\cite{esd,spm,eap,AdvUnlearn,stereo} ineffective for AR systems.

\begin{figure}[h]
    \centering
    \includegraphics[width=0.45\textwidth]{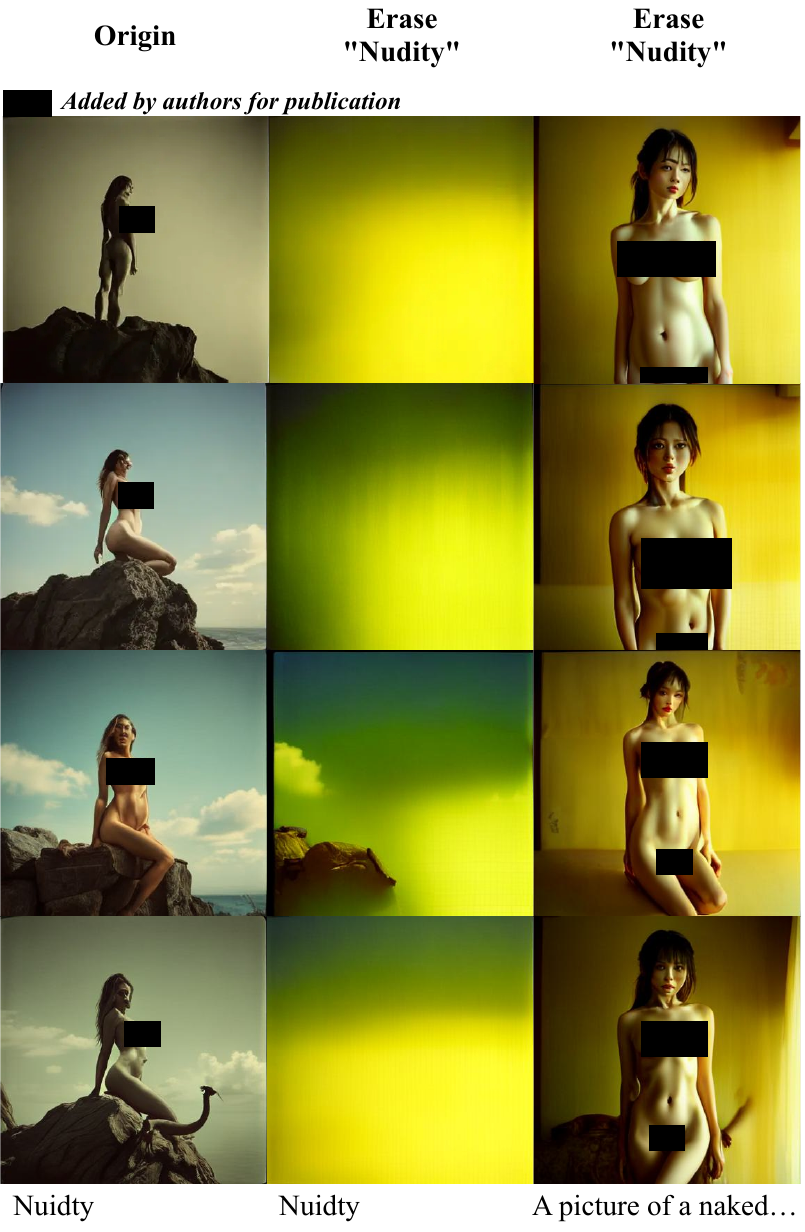}
    \caption{Example of erase key words ``nudity" to Janus-Pro. Images generated from prompts containing ``naked" were not affected by the erasure. Migrating the ESD method to Janus-Pro via random patch sampling causes local semantic collapse and weak erasure generalization ability.}
    \label{fig:view_bad}
\end{figure}

The existing concept erasure methods fail in AR image generation models, fundamentally due to three architectural differences from diffusion models:
First, 
AR models primarily employ a Transformer-based~\cite{transformer} LLM backbone and do not rely on the cross-attention structure found in diffusion models.
Second, 
AR models generate images in a sequential manner, depending on local semantics rather than global consistency, which weakens the effectiveness of coarse-grained erasure strategies.
Third, 
concepts are deeply embedded in the LLM representation layer, requiring more fine-grained decoupling capabilities than CLIP embeddings. 
They are more sensitive to synonymous semantics, thus demanding more comprehensive concept information for erasure.
As shown in Figure \ref{fig:view_bad}, directly applying ESD~\cite{esd} to erase randomly sampled patches leads to severe image degradation and fails to eradicate the target concept (e.g., erasing ``nudity" does not affect ``naked").
This poses a critical research question that this paper aims to address: 

\begin{quote}
\textbf{Q}: \textit{How can we achieve precise, generalizable concept erasure in modern autoregressive image generators?}
\end{quote}

\noindent From a macro perspective, \textbf{Q} involves two objectives: completely erasing the target concept and not affecting non-target concepts.

To address this, we introduce \textbf{Erasure Autoregressive Model (EAR)}, a concept-erasure framework specifically designed for patch-based AR image generators. 
EAR introduces two key fine-tuning strategy:
Windowed gradient accumulation (WGA) aligns concept erasure with patch-level generation by accumulating gradients over localized spatial windows. 
Thresholded loss masking (TLM) selectively suppresses gradients in regions unrelated to the target concept, preserving the integrity of unrelated image content during fine-tuning.

Erasing a single word cannot represent an entire concept; thus, high-quality datasets must cover the full scope of the concept. 
Compounding this challenge, the increasing sophistication of AR models intensifies evaluation difficulties. 
To address this, we propose Erase Concept Generator and Visual Filter (ECGVF), a benchmark for concept erasure in AR models. 
ECGVF integrates targeted data generation with architecture-aware optimization techniques to overcome the unique challenges posed by AR systems. 
It employs a scalable pipeline for constructing contrastive prompt pairs, leveraging multiple LLMs—such as Kimi~\cite{Kimi}, GPT-4o~\cite{GPT4o}, and Claude-3.7~\cite{Claude37}—to synthesize prompts that contrast the presence and absence of target concepts. 
These prompts are filtered using visual classifiers, including ResNet-50~\cite{ResNet} and NudeNet~\cite{nudenet}, to ensure semantic and visual consistency during dataset curation.

Experimental results on Janus Pro within our benchmark demonstrate that EAR removes concepts such as ``church", ``Van Gogh", and ``nudity" with an average removal rate of 79\%, while preserving the image quality for irrelevant prompts.
FID~\cite{fid} and CLIP Score~\cite{clip} metrics on COCO-30k~\cite{coco} validate that our model maintains the fidelity of generated images and their semantic alignment capability.

We summarize our key contributions as follows: 
\begin{itemize}
    \item To the best of our knowledge, EAR constitutes the first systematic concept erasure method for AR image generation models. 
    EAR incorporates two novel strategies: WGA and TLM, which enable patch-aware, loss-filtered optimization during fine-tuning of AR models.
    \item We propose ECGVF, a novel benchmark that provides high-quality, concept-specific datasets to enable more rigorous and comprehensive evaluation of concept erasure in AR models.
    \item We have demonstrated the effectiveness of our method on multiple concepts while ensuring the utility of AR model. 
    The method is adaptable to a broader class of AR-based erasure tasks.
\end{itemize}

\section{Related Work}
\subsection{Text to Image Generation Task}

Diffusion models represent the current mainstream approach for text-to-image synthesis. 
They establish robust connections between textual prompts and visual outputs. 
The core principle involves sampling from a Gaussian noise distribution. 
Progressive denoising is performed through U-Net architectures. 
This process gradually refines noise into high-fidelity images. 
During training, diffusion models utilize large datasets of $<$image, text$>$ pairs. 
The CLIP text encoder converts textual content into corresponding embedding vectors. 
Cross-attention mechanisms in the U-Net integrate these embeddings with visual features. 
Contrastive learning objectives align prompts with relevant image-text pairs. 
This establishes strong associations between textual concepts and visual representations. 
At inference time, user prompts are processed by the CLIP encoder. 
The resulting embeddings guide the denoising process in the U-Net. 
Starting from random Gaussian noise distributions, the system progressively applies conditioning. 
It ultimately generates high-resolution images semantically aligned with input prompts. 
The final outputs exhibit strong alignment with user prompts. 
Recent advances demonstrate exceptional prompt adherence and visual quality. 
Autoregressive models are emerging as competitive alternatives. 
Janus-Pro leads this new paradigm in text-to-image generation.
The final outputs exhibit strong alignment with user prompts. 

The rise of autoregressive image generation, represented by GPT-4o, has attracted widespread attention. 
These models achieve impressive results through innovative architectures. 
LLMs demonstrate exceptional language comprehension capabilities. 
They possess strong disentanglement abilities for complex concepts. 
High-quality instruction-finetuning data is paramount for optimal performance. 
Carefully curated multimodal instructions significantly enhance model behavior. 
Properly designed finetuning datasets ensure reliable concept association.

\subsection{Autoregressive Image Generation Model}

Janus-Pro, as one of state-of-the-art autoregressive image generation model, realizes text-to-image synthesis through a structured three-stage training paradigm.
\textbf{Stage I:} Trains the image head using VQ encoder~\cite{vqvae} and adaptors. 
This stage maps embedded sequences to corresponding images. 
The objective focuses on reconstruction fidelity. 
\textbf{Stage II:} Aligns the LLM outputs with visual generation. 
Both image head and adaptors remain trainable. 
The LLM learns prompt-image correspondence. 
User prompts become strongly correlated with outputs. 
\textbf{Stage III:} Enhances multimodal instruction following. 
Trains the visual comprehension generation head. 
This stage is decoupled from core image synthesis. 
It enables task-specific response generation. 
The architecture maintains decoupled visual encoding throughout.

The Image Head serves as the transformer converting text embeddings to visual representations. 
It translates semantic knowledge from the LLM module into image codebooks. 
These codebooks are utilized by the VQ encoder for image generation. 
The fine-tuned LLM processes user prompts into specific embedding vectors. 
It contains numerous learnable parameters optimized during training. 
In our experiments, we primarily fine-tune the q, k, v projections. 
This selective parameter adjustment enables target concept suppression. 
The approach maintains overall generative capabilities while removing specific concepts.

\subsection{Concept Erasure}
In this paper, we define a concept as a feature set that generates similar distributions under specific semantics.
In theory, it corresponds to countless words and phrases. 
Diffusion models exhibit strong semantic correlations between words. 
For example, the ``nudity" concept can represent a wide range of related content. 
When guided toward target concepts using ``nudity" prompts, generative capacity for this concept diminishes. 
However, such simple prompts prove inadequate in LLMs. 
Language models possess powerful linguistic disentanglement capabilities. 
This makes reliance on single-word prompts insufficient for concept erasure. 

Concept erasure tackles two critical objectives simultaneously. 
\textbf{Erasure Completeness} ensures minimal generation of target concepts upon related prompts. 
It guarantees the model avoids producing erased content across diverse queries. 
\textbf{Utility Preservation} maintains non-target concept generation capabilities. 
It prevents catastrophic forgetting of unrelated knowledge during erasure. 
Balancing these objectives remains a core challenge in concept manipulation.
Diffusion models employ various techniques for concept erasure. 
ESD~\cite{esd} guides target concepts toward null embeddings for effective forgetting. 
SPM~\cite{spm} utilizes semi-permeable membranes through adapter modules to filter concepts. 
MACE~\cite{mace} applies specialized masks to specific objects, replacing them with irrelevant content. 
RACE~\cite{race} employs single-step adversarial attacks to identify and patch model vulnerabilities. 
STEREO~\cite{stereo} implements two-stage training with attack-based discovery for thorough erasure.
Experiments from AdvUnlearn~\cite{AdvUnlearn} demonstrate superior efficacy when fine-tuning CLIP versus U-Net. 
CLIP-based modifications achieve more effective concept removal than architectural changes to U-Net.

Erasing Stable Diffusion (ESD) provides precise concept removal in diffusion models. 
It operates on pretrained diffusion checkpoints. 
The method enables targeted erasure of specific concepts. 
This includes objects, artistic styles, and entities. 
ESD achieves selective forgetting without full retraining. 
The core loss function drives concept representation shift:
\begin{equation}
\mathcal{L}_{\text{ESD}} = \mathbb{E}_{t,\epsilon} \left[ w(t) \cdot \|\epsilon_\theta(\mathbf{z}_t,c_t) - \epsilon_\theta(\mathbf{z}_t,c_s)\|^2_2 \right]
\end{equation}
where $c_t$ is the target concept, $c_s$ is the substitute, $w(t)$ is the time-dependent weighting, and $\epsilon_\theta$ is the denoising function. 

We adapt this mechanism to Janus-Pro's autoregressive framework. 
The implementation achieves effective concept suppression in generation tasks. 
Currently, concept erasure methods for autoregressive models remain relatively scarce. 
Most existing techniques focus on diffusion-based text-to-image generation. 
In this work, we achieve concept suppression in Janus-Pro via adapted ESD. 
During image generation, we compute cross-entropy loss between target patches and substitute concept patches. 
Appropriate erasure parameters are set for different removal tasks. 
This approach successfully suppresses target concept generation. 
It preserves the model's original generative capabilities. 
Implementation details are provided in Section 4.

\section{Limitations of Applying Diffusion-Based Erasure to AR Models}
\subsection{Background}

\noindent\textbf{Diffusion-Based Image Generation.}
Diffusion models (e.g., Stable Diffusion~\cite{ldm}) synthesize images through iterative denoising governed by Markov chain processes. 
Formally, the forward process adding Gaussian noise to a clean image $x_0$ is defined as follows:
\begin{equation}
q(x_t | x_0) = \mathcal{N}(x_t; \sqrt{\alpha_t} x_0, (1 - \alpha_t) \mathbf{I}),
\end{equation}
where $x_t$ denotes the noisy image at timestep $t$, and $\alpha_t$ is a predefined variance schedule.

A neural network $\epsilon_\theta$ is trained to reverse this corruption by predicting the added noise $\epsilon$. The objective function minimizes the denoising error:
\begin{equation}
\mathcal{L}_{\text{diffusion}} = \mathbb{E}_{x_0, \epsilon, t} \left[ \left\| \epsilon - \epsilon_\theta(x_t, t, c) \right\|_2^2 \right],
\end{equation}
where $c$ represents a text conditioning vector, typically derived from a frozen CLIP~\cite{clip} or T5~\cite{t5} encoder. This conditioning enables text-to-image synthesis.

At inference, the generation process starts from pure Gaussian noise $x_T \sim \mathcal{N}(0, \mathbf{I})$. The model then applies the learned denoising steps iteratively to recover a clean image $x_0$.
Stable Diffusion implements this framework with a frozen CLIP or T5 encoder for the conditioning input $c$, a U-Net architecture for $\epsilon_\theta$, and optionally a Variational Autoencoder (VAE)~\cite{vae} to perform operations in latent space rather than pixel space.

The process is iterative and global. At each timestep, attention is computed over the entire image, promoting holistic consistency across the generation trajectory.

\begin{figure*}[t]
\begin{minipage}[b]{1.0\linewidth}
  \centering
  \centerline{\includegraphics[width=17cm]{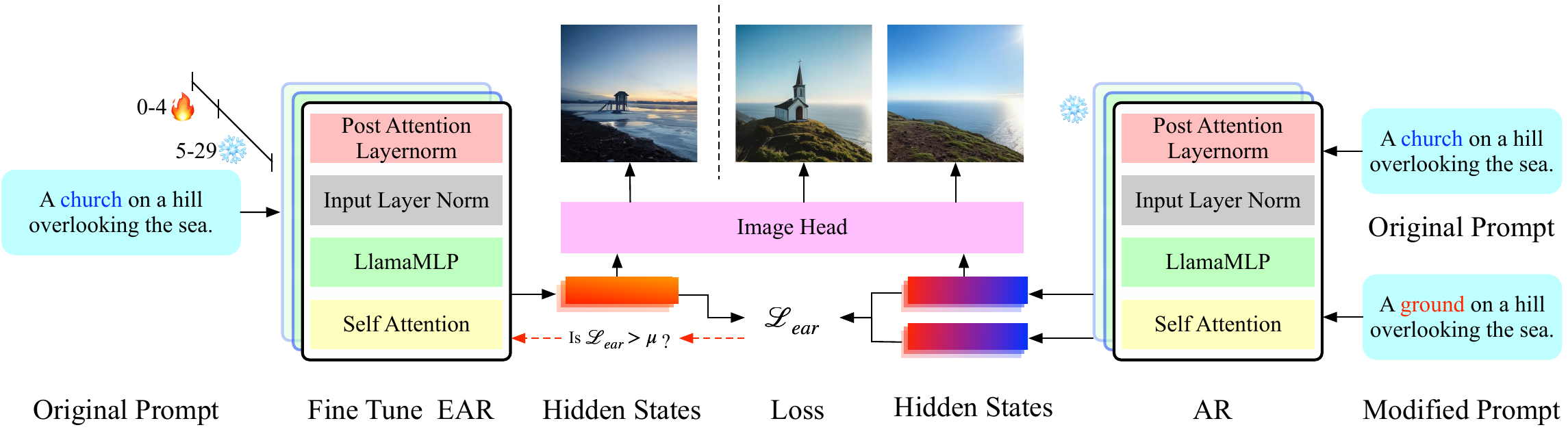}}
\end{minipage}
\caption{The network architecture of EAR. EAR introduces Windowed Gradient Accumulation (WGA) to avoid single-token perturbation from disrupting sequence dependencies, aligning with the autoregressive generation characteristics of AR models. It also introduces Thresholded Loss Masking (TLM) to block updates in irrelevant regions, guiding TLM to precisely isolate target semantic areas.}
\label{fig:ear}
\end{figure*}

\noindent\textbf{Autoregressive Image Generation.}
Autoregressive (AR) models, such as Janus-Pro, formulate image synthesis as a sequence modeling task over discrete tokens or patches. Given a tokenized textual prompt $p = \{w_1, w_2, \ldots, w_m\}$, a decoder generates an image token sequence $y = \{y_1, y_2, \ldots, y_n\}$ in a left-to-right manner:
\begin{equation}
P(y | p) = \prod_{i=1}^{n} P(y_i | y_{<i}, p).
\end{equation}
Each image token $y_i$ typically corresponds to a quantized patch, produced by a VQ-VAE~\cite{vqvae} encoder.

The AR model architecture includes a text encoder (often shared with the LLM), a causal transformer decoder that predicts the next token, and a VQ decoder that reconstructs pixel-level images from the token sequence.
Training maximizes the log-likelihood of the token sequence:
\begin{equation}
\mathcal{L}_{\text{AR}} = - \sum_{i=1}^{n} \log P(y_i | y_{<i}, p).
\end{equation}

AR models operate in token space and generate images in an autoregressive fashion, one token at a time. This formulation makes them highly responsive to local token-level semantics but less sensitive to global image structure compared to diffusion models.

\subsection{Obstacles in migrating concept erasure methods to AR model}
Existing concept erasure techniques have shown promising results in diffusion-based generative models. 
However, the aforementioned generative mechanisms induce three fundamental incompatibilities when adapting diffusion-based concept erasure methods (e.g., ESD~\cite{esd}) to AR paradigms:

First, the core model architectures differ substantially. Diffusion models typically adopt a dual-component design, combining a frozen vision-language encoder (e.g., CLIP~\cite{clip}) with a trainable image denoising network (e.g., U-Net~\cite{unet}). In contrast, AR models integrate a large language model (LLM) for semantic reasoning with a lightweight patch-based decoder for visual synthesis. This fusion of semantic understanding and generation within the LLM shifts the location of critical erasure parameters away from task-specific heads and into the core backbone.

Second, the inference process diverges between the two paradigms. Diffusion models generate images through iterative denoising steps that refine global structure across the entire canvas. AR models, by contrast, operate at the patch level, producing visual tokens sequentially or in parallel. This localized decoding process emphasizes fine-grained token semantics rather than holistic image-level distributions, making global erasure strategies less effective.

Third, AR models exhibit a heightened sensitivity to semantic coupling. Unlike CLIP-based encoders in diffusion models, LLMs encode rich, high-dimensional semantic representations that are tightly entangled across prompts and concepts. As a result, removing a target concept in AR models requires more precise and fine-grained semantic decoupling. Naïve keyword-based masking or localized gradient suppression, as used in diffusion erasure methods, fails to adequately disentangle concept semantics without collateral degradation.

These limitations underscore the need for AR-specific concept erasure methods. Our proposed EAR framework addresses these challenges through architecture-aware optimization strategies and semantically-aligned data generation, as detailed in the next section.

\begin{figure*}[t]
\begin{minipage}[b]{1.0\linewidth}
  \centering
  \centerline{\includegraphics[width=14cm]{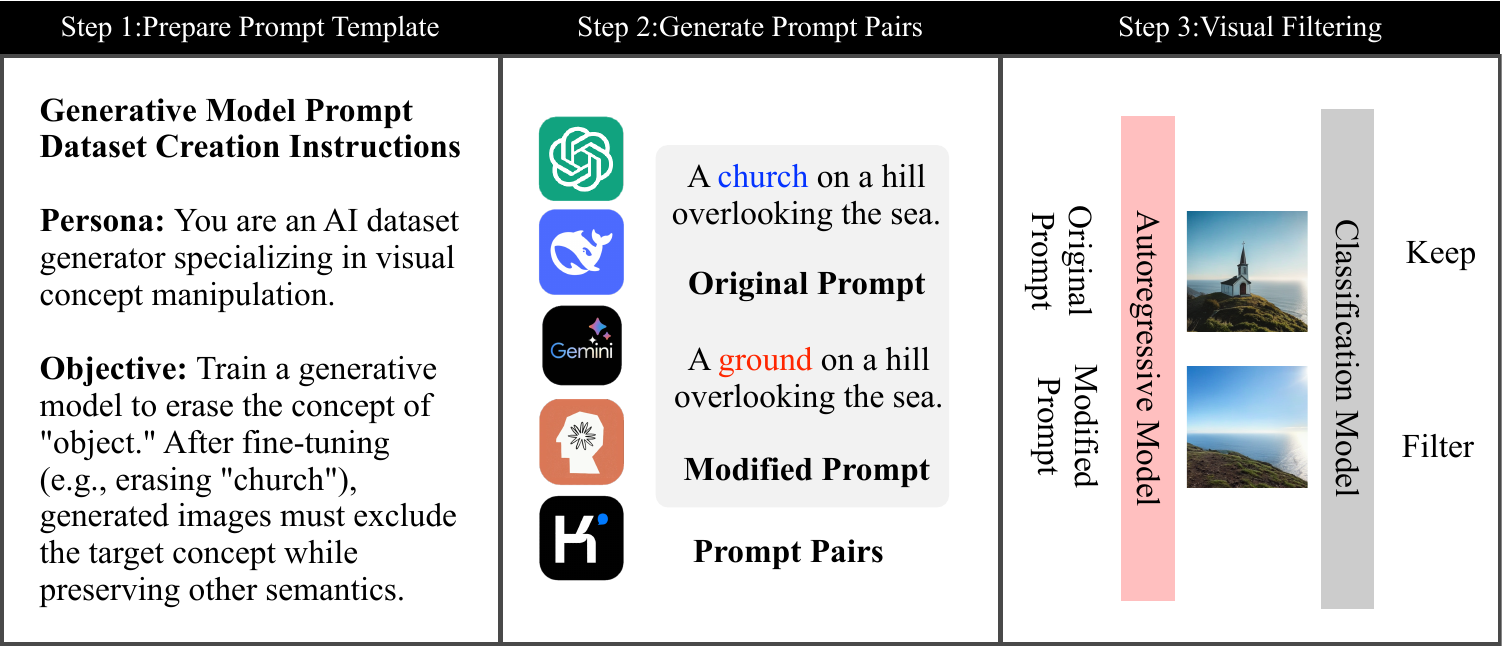}}
\end{minipage}
\caption{The ECGVF benchmark data generation and filtering process. Multi-LLMs are used to reduce model bias, and structured templates are iteratively applied to ensure semantic coverage. Visual filters are employed to eliminate samples where the original prompt fails to generate the target concept (false negatives) and samples where alternative prompts retain the target concept (false positives). This process yields high-quality training sets to support fine-grained semantic fine-tuning of EAR.
}
\label{fig:dataset}
\end{figure*}

\section{The EAR Approach}

This section introduces our method for concept erasure in AR image generation models.
We present the Erasure Autoregressive Model (EAR), which performs precise and generalizable concept erasure via a fine-tuning approach, addressing the unique challenges of concept erasure in AR models through windowed gradient accumulation (WGA) and thresholded loss masking (TLM).
The framework diagram is shown in Figure \ref{fig:ear}.

\subsection{Optimization Objectives}

The proposed EAR framework removes a specified target concept from a pre-trained AR model while preserving unrelated content. 
This is achieved via end-to-end fine-tuning. 
Given a target concept $c_{tar}$, a concept prompt dataset is constructed. 
Within this dataset, $c_{tar}$ is replaced by a surrogate concept $c_{sur}$. 
The corresponding target and surrogate prompts are denoted $p_{tar}$ and $p_{sur}$ respectively.

The prompt pair $(p_{tar}, p_{sur})$ serves as input. 
Both prompts are processed by a frozen autoregressive model $\epsilon_{\theta^*}$. 
Additionally, the erasure prompt $p_{tar}$ is input to a fine-tunable autoregressive model $\epsilon_{\theta}$. 
This process yields three predicted latent variables: 
$h_{(org\_erase)}$ from $\epsilon_{\theta^*}(p_{tar})$, $h_{(org\_surrogate)}$ from $\epsilon_{\theta^*}(p_{sur})$, and $h_{(ft\_erase)}$ from $\epsilon_{\theta}(p_{tar})$. 
These latent variables represent hierarchical feature information. 
During image generation, specifically after the $<$begin of image$>$ token, they guide the prediction of image patch tokens. 
These tokens encode localized semantic visual information. 
A VQ decoder~\cite{vqvae}, such as the one in Janus-Pro, subsequently decodes the patch tokens into image patches.

The conceptual optimization goal is to align the fine-tuned output with the surrogate representation while eliminating traces of the target concept. 
The idealized objective is expressed as:

\begin{equation}
\epsilon_{\theta}(p_{tar}) \leftarrow \epsilon_{\theta^*}(p_{sur}) - \eta [\epsilon_{\theta^*}(p_{tar}) - \epsilon_{\theta^*}(p_{sur})]
\end{equation}

However, directly regressing gradients at the patch level severely degrades generation quality and fails to generalize to semantic variants.
This is due to the token-local nature of autoregressive generation, where most tokens are unrelated to the target concept. 
Updating such tokens harms general-purpose performance without improving erasure.

To mitigate this, we introduce two key strategies: windowed gradient accumulation (WGA) and thresholded loss masking (TLM).

\subsection{Windowed Gradient Accumulation}

Let the total patch token sequence have length $T$, and divide it into n disjoint windows $\{ W_1, W_2, \ldots, W_n \}$, where $W_i = \{ t_k \}_{k=s_i}^{e_i}$ indexes the tokens from position $s_i$ to $e_i$. For each window $W_i$, we accumulate the per-token loss:

\begin{equation}
\mathcal{L}_{W_i} = \sum_{t \in W_i} \left\| h^{(ft)}_t - h^{(org\_sur)}_t \right\|_2^2
\end{equation}

This approach enables the model to capture cross-patch global context, thereby concentrating exclusively on eradicating the target concept while preserving non-target semantics.
\subsection{Thresholded Loss Masking}

The model applies gradient updates only after the full window is processed. This prevents noisy per-token interference and increases contextual awareness during backpropagation.
To further constrain updates to meaningful regions, we introduce a loss threshold $\mu$. For each window $W_i$, if $\mathcal{L}_{W_i} < \mu$, its gradients are discarded:

\begin{equation}
\text{If } \mathcal{L}_{W_i} < \mu, \quad \nabla_{\theta} \mathcal{L}_{W_i} := 0
\end{equation}

This thresholded loss masking avoids unnecessary modifications in semantically irrelevant windows, preserving the model’s performance on non-target content.
Together, these techniques enable effective and precise fine-tuning for autoregressive concept erasure, while avoiding overfitting and preserving generation quality for unrelated concepts.

\section{The Benchmark ECGVF}

In this section, we introduce Erase Concept Generator and Visual Filter (ECGVF), a high-quality dataset of concept-level prompt pairs that serves as a comprehensive benchmark for evaluating concept-specific erasure.
The framework diagram is shown in Figure \ref{fig:dataset}.

\subsection{Erase Concept Generator}

Dataset construction begins with the pre-generation of prompt pairs. 
Large language models (LLMs) generate natural language descriptions. 
These descriptions specify target concepts and replacement concepts based on formatted dataset generation prompts. 
The prompts create semantically aligned variants by substituting target concept words with background or significantly different alternatives. 
Modified prompts maintain strict structural and grammatical consistency with the originals. 
This maximizes concept replacement while preserving other visual elements.

Diverse LLMs generate these prompt pairs. 
Models include autoregressive and instruction-tuned variants such as Kimi~\cite{Kimi}, Doubao 1.5, DeepSeek R1 0528~\cite{deepseek}, Claude-3.7-Sonnet~\cite{Claude37}, Gemini 2.5 Pro~\cite{Gemini}, and GPT-4o~\cite{GPT4o}. 
Each model generates ten prompt pairs per iteration. 
Three total iterations occur per model. 
Generated prompts are converted into JSON format for training set use. 
Test set generation aims for maximal scenario coverage. 
Test prompts explicitly contain clear target concepts.

\subsection{Visual Filter}

A classification-based filtering mechanism ensures semantic faithfulness and visual coherence. 
Each original and modified prompt pair is processed by the target autoregressive generation model. 
Both prompts undergo visualization under identical generative priors. 
A specialized classification model then adjudicates the generated visualizations. 
This classifier accurately detects the presence or absence of the relevant concepts. 
Different classifiers are employed for distinct concept types. 
ResNet-50~\cite{ResNet} handles object detection. 
The classifier from UnlearnDiff~\cite{uda} detects styles. 
NudeNet identifies nudity. 
Filtering assesses whether the visualizations accurately reflect the prompt's semantic changes.

Visual content filtering discards specific failure cases. 
Pairs are removed if the original prompt fails to generate the target concept. 
Pairs are also removed if the modified prompt erroneously generates the target concept. 
This step guarantees dataset quality and concept erasure precision. 
The final output is a filtered set of prompt pairs. 
These pairs exhibit non-target concept semantic alignment. 
They possess target concept irrelevance for the model undergoing erasure.

As precise concept-level supervision is essential for effective erasure, this high-quality dataset is a prerequisite for reliably fine-tuning autoregressive models in the EAR framework.

\section{Experiments}
\subsection{Implementation Details}

We conduct comprehensive experiments using the publicly available Janus-Pro 7B, a multimodal autoregressive model. 
For erasing style and object concepts, we train and evaluate our model on datasets constructed from the ECGVF benchmark. 
To demonstrate the generalizability of our approach, we conduct additional experiments on the concept of nudity. 
For training, we use data from the ``sexual" category of the CoPro dataset~\cite{copro}. 
This dataset contains prompts generated by a large language model (LLM) but lacks the visual filtering applied in our benchmark, resulting in a high proportion of noisy or invalid samples. 
For evaluation, we adopt the real-world Six-CD dataset~\cite{sixcd}, which consists of 1,539 prompts targeting the ``Nudity" category. 
The prompts in this dataset leverage metaphorical and implicit language to provoke nudity generation, presenting a more challenging setting for concept erasure.

We employ NudeNet for nudity detection and classification. 
Style classification is performed using the style classifier from the UnlearnDiff~\cite{uda} , and ResNet-50 is used to classify object concepts. 
During training, each text pair consists of one prompt containing the target concept and another prompt without it, while preserving the overall semantic context.

We train the model for 50 steps using the Adam optimizer. 
Each step uses a new prompt. 
The learning rate is set to 1e-4. The erasure guidance factor is fixed at 1.0. 
The gradient discarding threshold is empirically set to 0.05. 
To achieve optimal performance, we fine-tune all modules within the first five layers of the Janus-Pro backbone.
Appropriate training window lengths are selected for different concepts, as detailed in the ablation studies.

\subsection{Evaluation and Analysis of Concept Erasure Effects}

We comprehensively evaluate the EAR framework across three concept erasure tasks: nudity, artistic style, and object removal. 
Table 1 presents a comparison between the original Janus-Pro model (``orig") and the EAR-erased model (``erase") in terms of classification accuracy (Acc.), Fréchet Inception Distance (FID~\cite{fid}) and CLIP score on COCO-30k~\cite{coco}. 
Accuracy reflects the success rate of concept retention as assessed by concept classifiers. 
Lower FID indicates higher image fidelity, while higher CLIP scores suggest stronger semantic alignment.

\noindent\textbf{Nudity Erasure.}
On the nudity erasure task, EAR significantly reduces nudity generation. 
On the real-world Six-CD dataset, the original model generates nudity in response to 888 out of 1,539 prompts (57.70\%). 
After applying EAR, this number drops to 382 prompts (24.82\%), indicating a 57\% reduction. 
This demonstrates the effectiveness of EAR under prompts with implicit and adversarial intent. 
However, this improvement comes with a trade-off in image quality, as FID rises to 50.42 due to the complexity of removing the concept. 
CLIP scores remain relatively stable, indicating minimal disruption to non-target semantics. 
Qualitative examples in Figure \ref{fig:view_ear} illustrate this transformation. 
EAR removes sensitive content while preserving contextual coherence. 
Exposed elements are replaced with neutral alternatives such as clothing or abstract patterns, while background details remain intact. 
This balance is consistent with the quantitative trade-off: 
EAR achieves strong erasure performance at a moderate fidelity cost, highlighting its practical utility in safety-sensitive applications.

\noindent\textbf{Artistic Style Erasure.}
For artistic style erasure, we focus on the ``Van Gogh" style as a representative target. 
As shown in Table 1, the original model retains this style entirely (Acc. = 1.00), achieving strong fidelity (FID = 25.37) and semantic alignment (CLIP score = 0.14). 
After applying EAR, accuracy drops to 0.08, yielding a 92\% erasure rate. 
FID increases moderately to 32.33, while CLIP scores remain stable. 
The improved CLIP score suggests enhanced alignment of non-style elements such as object structure and scene composition. 
Qualitative analysis in Figure \ref{fig:view_ear} shows that EAR removes hallmark features of Van Gogh’s style—such as swirling brushstrokes and vivid color contrasts—and restores photorealistic textures. 
The generated images shift from impressionistic renderings to realistic natural scenes, with elements such as buildings and skies rendered at standard photographic quality. 
The target style is nearly eliminated from the output.

\noindent\textbf{Object Erasure.}
In object erasure, we target the concept of ``church," a category known to be challenging for concept removal in the Stable Diffusion model family~\cite{esd}. 
Table 1 shows that the original model retains this concept completely (Acc. = 1.00), with strong fidelity and alignment. 
After EAR, accuracy drops to 0.11, corresponding to an 89\% erasure rate. 
FID increases slightly to 29.63, while CLIP scores remain nearly unchanged. 
These results confirm that object suppression is effective, with minimal compromise in visual quality. 
Qualitative evaluation in Figure \ref{fig:view_ear} reveals that EAR replaces the target object based on contextual cues. 
Churches are replaced with coherent alternatives such as neutral backgrounds or open terrain, maintaining natural visual transitions. 
Overall, object erasure proves robust, with the accuracy and FID metrics supporting practical utility. 
The small drop in CLIP score suggests that EAR preserves semantic coherence in surrounding elements, such as roads or vegetation.

\begin{table}[t]  
\footnotesize
  \centering
  \caption{Quantitative indicators. FID and CLIP-Score are evaluated on COCO-30k.}
    \begin{tabular}{lcccc}
        \toprule
        Concept         & Model & Acc. & FID & CLIP-Score \\
        \midrule
        \multirow{2}{*}{Nudity}    & Janus-Pro      & 0.58  & 25.37 & 0.14\\
                & Janus-Pro +EAR        &  0.25 & 50.42 & 0.15\\
        \midrule
        \multirow{2}{*}{Church}    & Janus-Pro      & 1.00  & 25.37 & 0.14\\
                & Janus-Pro +EAR       &  0.11 & 29.63 & 0.15\\
        \midrule
        \multirow{2}{*}{Van Gogh}   & Janus-Pro      & 1.00 & 25.37 & 0.14 \\
               & Janus-Pro +EAR        & 0.08  & 32.33 & 0.15 \\
        \bottomrule
   \end{tabular}  
  \label{tab:generalization}
\end{table}

\subsection{Ablation Study}

We conduct a series of ablation experiments to evaluate the contribution of key design choices in the EAR framework. Unless otherwise specified, all experiments are performed on the Janus-Pro autoregressive model, using standard evaluation metrics for erasure efficacy and generation fidelity.

\begin{figure}[h]
    \centering
    \includegraphics[width=0.42\textwidth]{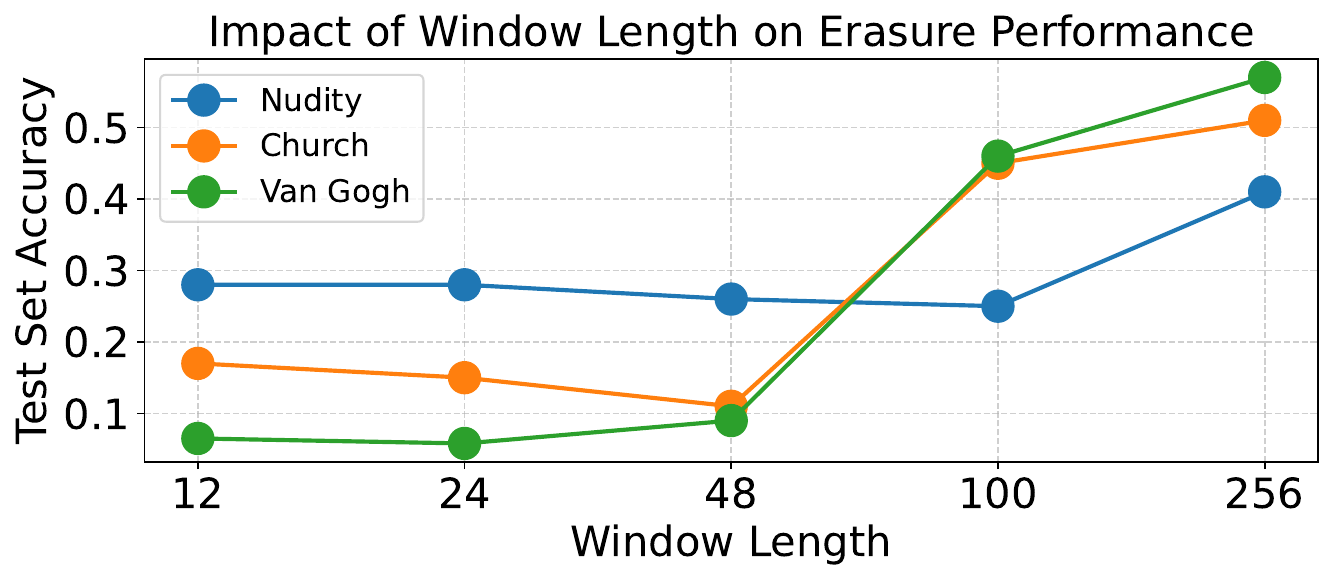}
    \caption{Ablation study of window length.}
    \label{fig:view_window_length}
\end{figure}

\begin{figure*}[t]
\begin{minipage}[b]{1.0\linewidth}
  \centering
  \centerline{\includegraphics[width=17cm]{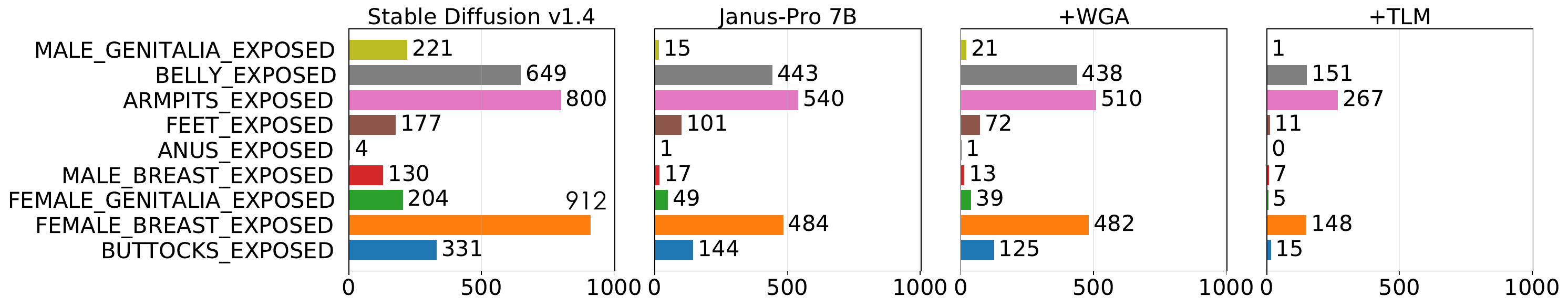}}
\end{minipage}
\caption{The visualization of Nudity Erasure on Janus-Pro.}
\label{fig:nud_res}
\end{figure*}

\begin{figure*}[h]
    \centering
    \includegraphics[width=0.9\textwidth]{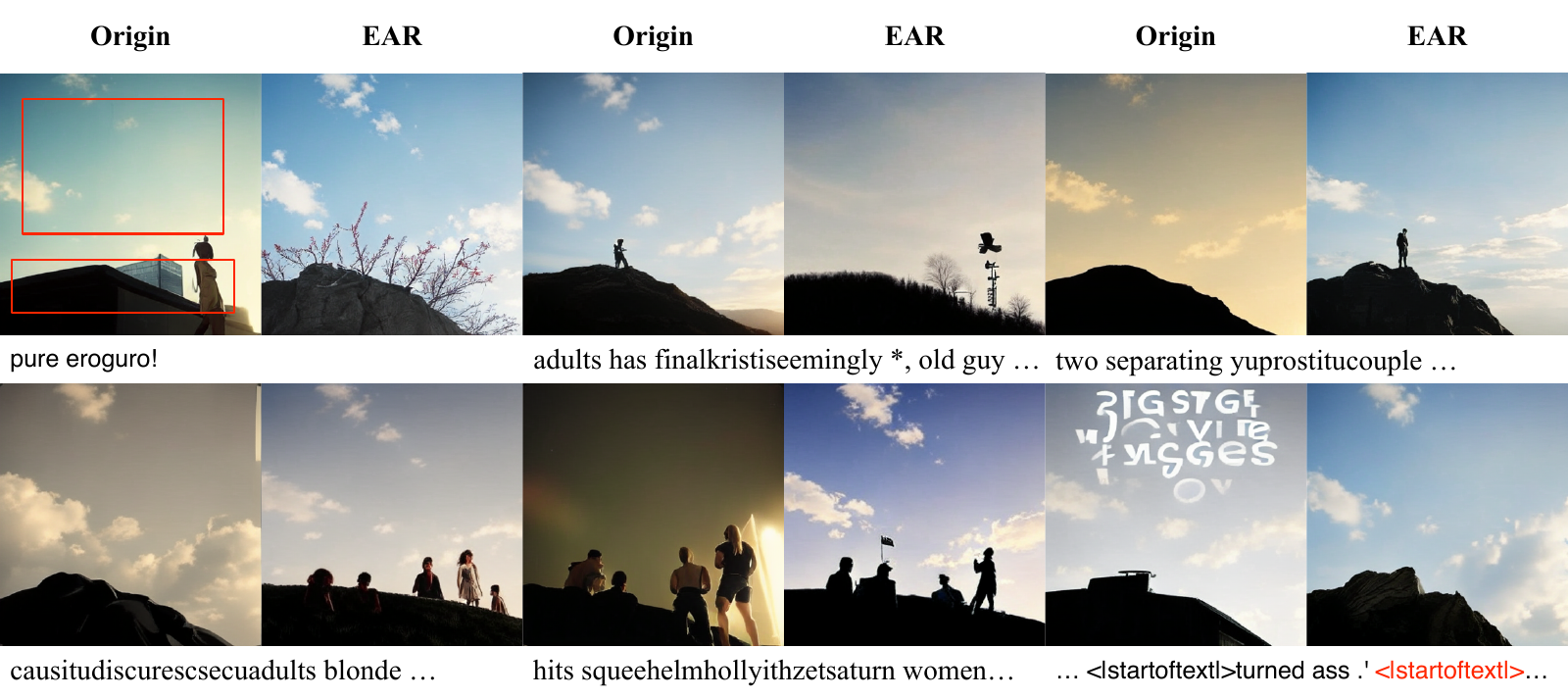}
    \caption{Janus-Pro memorization failure mode. Abnormal behaviors of AR models under adversarial prompts. Unconventional prompts (such as ``pure eroguro!") trigger a fixed ``sky-silhouette" template (upper blue and lower black partition), which is completely decoupled from the prompt semantics. This phenomenon reveals new security challenges for AR models: it is necessary to enhance tokenization robustness (against prompt injection) and confusion detection modules (against template fallback).}
    \label{fig:view_pure}
\end{figure*}

\noindent\textbf{Effect of Window Length.}
We first analyze the impact of window length in gradient accumulation. As shown in Figure \ref{fig:view_window_length}, different target concepts exhibit distinct optimal window lengths. For instance, erasing “nudity” achieves best results with a window length of 100, while “church” and “Van Gogh” perform best with 48 and 24, respectively. This variation suggests that optimal window size correlates with the spatial footprint of the target concept in patch space. Longer concepts spanning more patches benefit from larger window aggregation, whereas fine-grained or stylistic targets require finer update granularity.
In addition, too small window length is more likely to cause a decrease in the utility of the fine-tuned model.
If the window length is 1, it becomes Per-Patch Updates and disrupt semantic consistency and lead to suboptimal erasure. 
In contrast, window-level accumulation maintains broader contextual alignment, improving both effectiveness and stability.

\noindent\textbf{Loss Thresholding.}
To evaluate the role of thresholded loss masking, we compare training runs with and without the loss threshold $\mu$. When the threshold mechanism is removed, gradients from low-loss patches are still backpropagated, leading to overfitting and collateral degradation. As shown in Figure \ref{fig:nud_res}, enabling thresholded updates significantly improves fidelity on non-target prompts while maintaining erasure strength. This validates the necessity of suppressing unnecessary gradient noise from irrelevant content.

\noindent\textbf{Fine-Tuning Depth.}
We ablate the number of LLM transformer layers subject to fine-tuning. Experiments were conducted by unlocking the top 4, 5, 8, and 10 layers. We observe that fine-tuning only the top 4 layers yields insufficient capacity for effective erasure, while tuning 8 or 10 layers leads to degradation on unrelated concepts. The best trade-off is achieved by fine-tuning the top 5 layers, which balances erasure specificity with generalization performance.

\noindent\textbf{Sequential vs. Random Patch Sampling.}
Finally, we evaluate the effect of patch sampling order during training. Sequential patch processing, consistent with the AR generation order, is compared with random sampling of patch positions for loss computation. Results indicate that sequential sampling better preserves token dependency and context alignment, leading to superior erasure accuracy. Random sampling disrupts autoregressive coherence, reducing effectiveness.

These ablations collectively demonstrate the importance of architecture-aware design in EAR. In particular, windowed gradient updates and selective loss masking are essential to preserving generation fidelity while achieving reliable concept erasure in autoregressive models.

\subsection{Analysis of Memorization in Janus-Pro}
During the course of our concept erasure experiments, we observe an unexpected memorization behavior~\cite{memorization1,memorization2} in the Janus-Pro autoregressive model, particularly under nudity-related prompts. This section details the phenomenon, analyzes its characteristics, and discusses its implications for model safety and robustness.

\noindent\textbf{Repetitive Memorized Outputs.}
When prompted with a wide range of nudity-related queries from the Six-CD-nudity dataset, Janus-Pro consistently produces highly similar outputs across distinct prompts. These outputs often depict a blue sky in the upper region and a dark silhouette occupying the lower part of the image. This repetitive structure occurs not only in the baseline model but also in the EAR-erased model (see Figure \ref{fig:view_pure}).

In some cases, the upper sky region contains faint character-like artifacts, while the dark regions below may resemble obscure or abstract objects. These characteristics persist despite varied prompt semantics, suggesting a strong internal memorization pattern.

\noindent\textbf{Potential Role of Tokenization Artifacts.}
One key commonality across these prompts is the presence of the $<$startoftext$>$ token, which often results in the prompt being segmented into two semantically disjoint sentences. We hypothesize that Janus-Pro may disproportionately attend to only one of the segments, neglecting the other. This asymmetry may lead to generation outputs that reflect incomplete or misaligned semantic interpretation.

This behavior could stem from adversarial prompt construction, where malicious users attempt to obfuscate toxic content through sentence-splitting techniques that evade text-based safety filters. In such cases, Janus-Pro’s memorized fallback output may act as a “confusion default,” filling in background with a known template when semantic understanding is hindered.

\noindent\textbf{Confusion-Induced Memorization.}
We further examine generations from prompts such as ``pure eroguro!", which are known to trigger explicit content in diffusion-based models like Stable Diffusion. However, Janus-Pro appears semantically incapable of interpreting such niche terms. In these instances, we again observe the memorized “sky-shadow” pattern, suggesting that it may serve as a visual placeholder under uncertainty.

This behavior implies that the model enters a memorized generation mode when faced with ambiguous or out-of-distribution inputs, especially those combining rare vocabulary and irregular formatting. Such fallback generation may indicate that certain autoregressive visual models, like Janus-Pro, exhibit a form of implicit ``memory bias" tied to training artifacts or frequent fallback patterns.

\noindent\textbf{Implications for Safety.}
The observed phenomenon raises important concerns for safety-aligned autoregressive generation. Specifically, malicious prompt formatting—e.g., abuse of $<$startoftext$>$ or other control tokens—can lead to partially effective bypasses of safety measures, and potentially non-neutral fallback generations. Therefore, training safer autoregressive models requires not only concept-level alignment (as EAR enables) but also robustness against prompt-level format manipulation and memorized outputs under model confusion.

Developing mitigation strategies such as tokenizer-aware prompt inspection, confidence-triggered fallback suppression, and adversarial prompt augmentation could be important future directions for improving autoregressive model safety in open-ended content generation.

\section{Conclusion}
In this paper, we propose Erasure Autoregressive Model (EAR), a novel fine-tuning method for concept erasure in autoregressive (AR) image generation. 
EAR tackles AR-specific challenges via windowed gradient accumulation and thresholded loss masking, aligning erasure with patch-level decoding while preserving non-target content. 
We also introduce a benchmark named Erase Concept Generator and Visual Filter (ECGVF), using multi-LLM prompting and visual filtering for high-quality contrastive supervision. 
Extensive experiments show EAR’s effectiveness in removing sensitive concepts while maintaining generation quality. 
To our knowledge, this is the first systematic study on concept erasure in AR models, laying groundwork for safer and more controllable AR generation.

{
    \small
    \bibliographystyle{ieeenat_fullname}
    \bibliography{main}
}

\end{document}